# Hand Gesture Recognition Based on Karhunen-Loeve Transform

Joyeeta Singha[1], Karen Das[2]

[1,2]Department of Electronics and Communication Engineering
[1,2]Assam Don Bosco University, Guwahati, Assam, India

*Abstract:* In this paper, we have proposed a system based on K-L Transform to recognize different hand gestures. The system consists of five steps: skin filtering, palm cropping, edge detection, feature extraction, and classification. Firstly the hand is detected using skin filtering and palm cropping was performed to extract out only the palm portion of the hand. The extracted image was then processed using the Canny Edge Detection technique to extract the outline images of palm. After palm extraction, the features of hand were extracted using K-L Transform technique and finally the input gesture was recognized using proper classifier. In our system, we have tested for 10 different hand gestures, and recognizing rate obtained was 96%. Hence we propose an easy approach to recognize different hand gestures.

## 1. INTRODUCTION

Hand gesture recognition is one of the growing fields of research today which provides a natural way of human-machine interaction. Gestures are some forms of actions which a person expresses in order to express information to others without saying it. In our daily life, we can observe few hand gestures frequently used for communication purpose like thumbs up, thumbs down, victory, directions etc. Some common examples are in cricket where the umpire uses different hand gestures to show different events that occurred at that instant on the match, hand gestures used by the traffic police, etc.

In our paper, we have firstly used Skin filtering where the RGB image is converted to HSV image because this model is more sensitive to changes in lighting condition. And then K-L transform is performed. The advantage of K-L transform is it can eliminate the correlated data, reduces dimensions keeping average square error minimum, and gives excellent cluster character after the transform. Some applications in this field that has already been done, for example hand gesture recognition for sign language, hand gestures used for controlling robot's motion, in video games, etc.

## 2. LITERATURE REVIEW

We have studied many previous works done in this field by different researchers. There are many approaches that were followed by different researchers like vision based, data glove based, Artificial Neural Network, Fuzzy Logic, Genetic Algorithm, Hidden Markov Model, Support Vector Machines etc. Some of the previous works are given below.

Many researchers [1][2][3][4][5][6][7][8][9][10][11] used Vision based approaches for identifying hand gestures. Kapuscinski [1] found out the skin colored region from the input image captured and then this image with desired hand region was intensity normalized and histogram was found out for the same. Feature extraction step was performed using Hit-Miss Transform and the gesture was recognized using Hidden Markov Model (HMM). Recognition rate obtained was 98%. Yu [2] used YCbCr colour model to distinguish skin coloured pixels from the background. The required portion of the hand was extracted using this colour model and filtered using median filter and smoothing filter. The edges were detected and features extracted were hand perimeter, aspect ratio, hand area after which Artificial Neural Network (ANN) was used as classifier to recognize a gesture. Accuracy rate obtained was 97.4%. In [3][8] fingertip detection was used for hand gesture recognition. Rajam [3] in his paper for sign language recognition first converted the RGB image captured to binary and Canny Edge Detection Technique was used for extracting edge of the palm. The fingertip positions of the fingers were identified from the extracted edge of palm by measuring their distance from a reference point which is taken to be at the bottom of the palm. Recognition rate obtained was 98.125%. Raheja [8] scanned the skin filtered image in all direction to find out the edges of the fingers and the tips of the edges were assigned the highest pixel value and as such fingertip was detected. Malima [4] used hand gesture recognition for controlling the robot. The Red/Green ratio was found out which was used for determining the skin coloured regions. The centre of gravity of the hand was found out along with the farthest distance from it and thus in such a way the finger tips were determined. A circle was made around the centre of gravity and number of white pixels beyond that circle was counted to know the desired gesture. Recognition rate obtained was 91%. Jackin [5] used the same steps as [4] except that the RGB input was converted to HSV colour space before undergoing further steps. Almost 100% accuracy was obtained. Huang [6] found the skin coloured pixels from the image, after which the features like orientation, spatial frequency, spatial locality were extracted for which Gabor filter and Principle component analysis(PCA) was used. Support vector machines (SVM) was used as classifier for this paper. Recognition rate obtained was 95.2%. Koh [7] used an extra step at the beginning that is the Active Appearance model which considers the shape and colour of the image. This model finds out the rough contour of the hand. Recognition rate obtained was 82.6%. Rekha [9] converted input image to YCbCr skin colour model, using proper threshold method desired hand part was extracted from the input image. Principle Curvature Based Region detector (PCBR) and 2-D Wavelet Packet Decomposition (WPD) techniques were used for feature extraction. Recognition Rate obtained was 91.3% for static and 86.3% for dynamic. Gopalan [10] found the skin coloured pixels and regions corresponding to such pixels were cropped out. The orientation of the image was found using PCA. Features extracted were distance, angle by which each points on the contour was related to one another by IDSC (Inner distance shape context) algorithm and finally gesture was recognized by SVM. Fang [11] applied Adaptive Boost algorithm for detecting hand from the input image. The main advantage of using this algorithm is, it not only could detect a single hand but also could detect overlapped hand. Features extracted were palm and finger structures which were determined by drawing blobs and ridges. Recognition rate obtained was 98%.

In [12][13][15][16] Data-Glove Based Approaches was used. Kumar [12] used DG5 VHand 2.0 data gloves for hand gesture recognition. The features like position of fingers, hand was given by the gloves after which KNN classifier algorithm was used. He used it in some applications like air writing and image browser. Kim [13] used KHU-1 data glove along with Kinematic chain theory to extract the features like joints from hand. Glove consisted of 3 accelerometer sensor, a controller and a Bluetooth. Finally, rule based algorithm was used for gesture identification. Weissmann [15] used Cyberglove that took into account angles made by 18 joints of hand. Features extracted using this glove were angle made between the neighbouring fingers, wrist pitch, thumb rotation which was then trained using ANN. Cavallo [16] used glove where 18 markers were attached with it, of which 15 were for fingers and 3 for the reference taken. The image captured was then classified based on Singular value decomposition (SVD).

In [17][18][19][24] ANN based system was proposed for recognizing the gesture. It was used because of advantageous factors like generality, adaptive learning, self-organizing and real time operations. Paulraj [17] used ANN to recognize American Sign Language. Firstly, the Skin coloured regions were extracted after which the moment invariant was obtained. ANN was used where the network has 58 neurons of which 34 were input neurons, 20 hidden neurons and 4 output neurons and the dataset comprised of 270 feature vectors. Admasu [18] used ANN for recognizing the Ethiopian Sign language of 34 letters. The neural network consisted of three layers, input layer comprising of 200 neurons giving the number of feature vectors, output layer with 34 neurons describing the number of categories which was to be recognized and hidden layer with 100 neurons. Backpropagation algorithm was used for training. Recognition rate achieved was 98.53%. Oniga [19] used ANN for recognizing the hand gesture used for intelligent human-machine interface. The neural network comprised of 2 layers with 25 neurons, the hidden layer with 20 neurons and the output layer with 5 neurons. It was processed and trained by Back propagation method with Levenberg-Marquardt algorithm along with 1500x5 vector dataset. Lamar [24] used ANN with back propagation algorithm which had 20 or 24 neurons in input layer, 42 neurons each in the hidden and output layer. The network after trained by an input vector gave a single output neuron giving the desired recognition.

In [20][21] Genetic Algorithms (GA) was used for solving problems in which steps were selection of parent data, recombination and mutation. The reason for its frequent use is it helps in getting optimal solutions to the problem. Ghotkar [20] used the above algorithm in recognizing Indian Sign Language and said that the randomness of the samples which was taken at the input can be managed properly using this approach but they were able to recognize only the few alphabets of Indian Sign Language which was one of the disadvantage. Lee [21] used GA to extract robust fingertip for interacting with the robot. YCbCr colour model was used but only to predict the probability of whether the pixels are skin coloured or not. After the prediction, a very important step comes that is the threshold selection which in this paper is done using GA. During this algorithm, initial population was 8 bits and function determining the fitness was achieved using Otsu's method.

Fuzzy logic is a problem solving approach based on degrees of truth rather than the usual true or false i.e. 1 or 0. It includes 0 and 1 as extreme cases of truth and also includes the various states of truth in between. For example intelligence cannot be measured with normal 1 or 0. It has to be compared with other's intelligence and result can be 0, 1 or in between. In [22][23] fuzzy logic was implemented for hand gesture recognition. Verma [22] said that the motion of the hand can be detected by Finite state machine's (FSM) states. These states are assumed as clusters which are indeed formed by fuzzy c-means clustering. Then the centroid of each clusters found out mathematically, and hence states of FSM was determined and finally gesture was recognized.

Kim [23] used Fuzzy logic for recognizing Korean Sign Language. With time, the position and speed of hand changes, these different speeds were considered to be the fuzzy sets which were indicated mathematically as zero, small, medium, large, etc.

In [24][25] Principal Component Analysis (PCA) was implemented in their works. Lamar [24] used PCA for extracting features from the input image in which mean, covariance, Eigen values and Eigen vector were found out. Mean describing the position of finger, Eigen value describing the shape of the finger, Eigen vector showing the direction of the image were the features used. Chang [25] in his paper used PCA as a classifier where Eigen face was extracted from the image to be tested and then the Euclidean distance was found between classes.

In [26][14] Hidden Markov Model (HMM) was used for hand gesture recognition in different fields. Singh [26] used HMM for identifying gesture while interacting with the robot. Features extracted like shape of hand, optical flow were fed to the HMM after which certain matrices like state transition matrices were obtained which helped in recognizing the gesture when an arbitrary hand was fed. Elmezain [14] used HMM to recognize alphabets while hand in motion. Features extracted in this paper were location, velocity whose vector was then fed to the HMM. Left-Right Banded model along with Baum-Welch algorithm was used for recognizing the gesture. Recognition rate obtained was 92.3%.

### 3. THEORITICAL BACKGROUND

*A. Canny Edge Detection*

This is one of the best edge detection techniques but little complex than other edge detection techniques. The major advantage of this technique is its performance. In case of other edge detection techniques only one threshold is used, in which all values below the threshold were set to 0. Thus, we must be very careful while selecting the threshold. Selecting the threshold too low may result in some false edges which are also known as false positives. Whereas if the threshold selected is too high, some valid edge points might be lost, this is also known as false negatives. But canny edge detection technique uses two thresholds: a lower threshold, TL and a higher threshold, TH thus eliminating problem of false positive and false negative. Steps involved in this type of detection are:

- The input image is smoothened with a Gaussian filter after which the Gradient magnitude and angle images are computed.
- Non-maxima suppression is applied to the gradient magnitude image.
- And finally detection and linking of the edges is done using double thresholding and connectivity analysis.

*B. Karhunen-Loeve (K-L) Transform*

K-L Transform is used to translate and rotate the axes and new coordinate is established according to the variance of the data. The K-L transformation is also known as the principal component transformation, the eigenvector transformation or the Hotelling transformation. The advantages are that it eliminates the correlated data, reduces dimension keeping average square error minimum and gives good cluster characteristics. K-L Transform gives very good energy compression. It establishes a new co-ordinate system whose origin will be at the centre of the object and the axis of the new co-ordinate system will be parallel to the directions of the Eigen vectors. It is often used to remove random noise. The steps involved in the process are:

- Firstly we consider an input data matrix say X and then we find out the mean vector say M
$$M = E\{X\}$$

- The next step is finding out the covariance matrix C of X.

Mathematically, covariance is given by
$$C = E\{(X - M)(X - M)'\}$$

- The eigen values and eigen vectors are found out from C such that the eigenvalues are arraged in the descending order and corresponding eigen vectors are obtained.
- Matrix A is obtained in such a manner that the first row represents eigen vector corresponding to maximum eigen value and so on.
- K-L Transform is given by:
$$KLT = A * (X - M)$$

where A is the matrix consisting of Eigen vectors arranged in rows such that they are arranged in decreasing order of Eigen value.

### 4. PROPOSED SYSTEM

The block diagram representing the proposed system is given in Fig. 1. The details of the system along with explanations and figures of each step are given below.

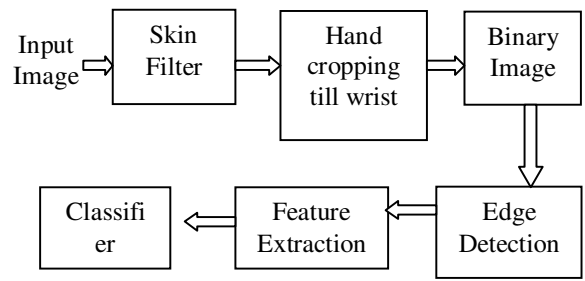

Fig. 1. Block diagram of the proposed approach

We have considered 10 hand gestures each having 10 samples. Some of the images we have experimented are shown in the fig. 2. Here, we have shown only 30 figures out of 100 figures.

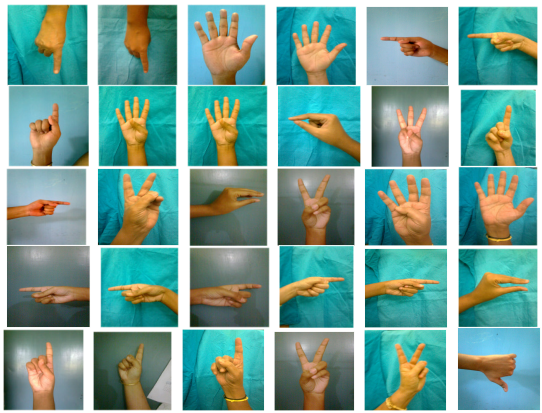

Fig 2. Some of the different images captured by camera

*B. Skin Filtering*

Skin filtering is a process of finding regions with skin-coloured pixels from the background. This process has been used for detection of face, hand, etc. which is applied in different fields. The basic block diagram of the skin filtering process is shown in Fig. 3.

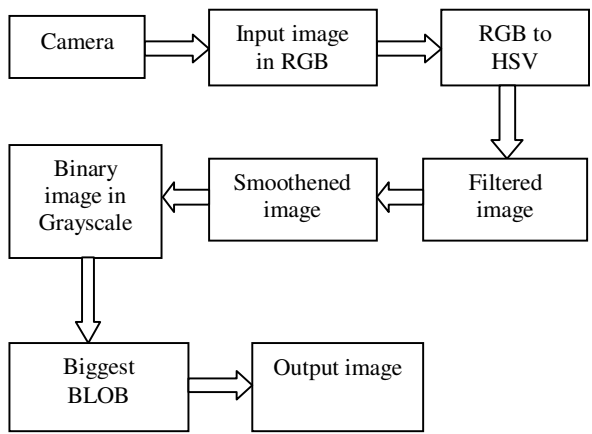

Fig. 3. Block diagram of skin filtering

The first step involves the capturing of image using camera and conversion of the input RGB image to HSV colour space. This step is done because HSV model is more sensitive to changes in lighting condition. HSV which means Hue (H), Saturation (S) and the brightness (I, V or L) is shown in Fig. 4.

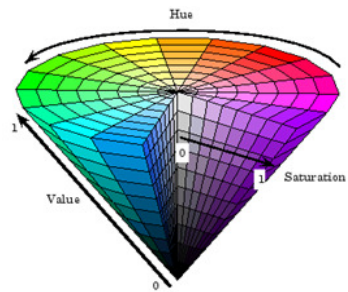

Fig. 4. HSV colour space

Hue describes the shade of the colour, saturation describes how pure the hue is with respect to white and brightness shows how much light is coming from the colour. Mathematically H, S and V values can be obtained from R, G and B values as:

$$H = \begin{cases} 60\left(\frac{G-B}{\delta}\right) & \text{if } MAX = R \\ 60\left(\frac{B-R}{\delta}+2\right) & \text{if } MAX = G \\ 60\left(\frac{R-G}{\delta}+4\right) & \text{if } MAX = B \\ \text{not defined} & \text{if } MAX = 0 \end{cases}$$

$$s = \begin{cases} \frac{\delta}{MAX} & \text{if } MAX \neq 0 \\ 0 & \text{if } MAX = 0 \end{cases}$$

where $\delta = (MAX - MIN)$, $MAX = \max(R, G, B)$, and $MIN = \min(R, G, B)$.

The resulting image was filtered, smoothened and finally we obtain a gray scale image. Along with the desired hand image, other objects having skin coloured was also taken into consideration which needs to get removed. This was done by taking the biggest BLOB (Binary linked object). An example showing the skin filtering steps is shown below in Fig. 5.

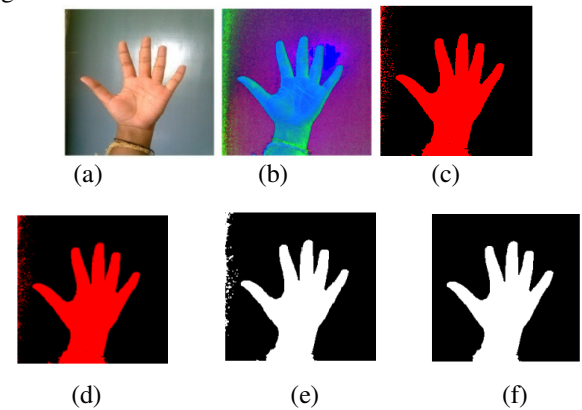

Fig. 5. Skin filtering: a) Input image, b) HSV conversion of input image, c) Filtered image, d) Smoothened image, e) Binary image in grayscale, f) Biggest BLOB.

## B. Hand Cropping

After the skin has been extracted from the input image, hand cropping is done. As we are considering the gesture shown by the hand only till the wrist portion, it is important to remove the other skin parts. Here, we scanned the image from left to right or vice-versa and from top to bottom or vice-versa in order to detect the wrist as explained in [8] and then the minimum and maximum positions where the white pixels end was determined and based on that cropping is done. One of the arbitrary images has been shown in Fig. 6.

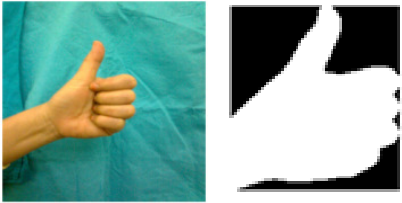

Fig. 6. Cropping of hand. (a) Input image, (b) cropped image

## C. Edge Detection

After extraction of the desired hand we extracted only the edges of the hand in the image so as to further reduce the time and computational complexity during the whole process. There are different edge detection techniques like Sobel, Prewitt, Roberts, Gaussian, zero-cross and canny method among which Canny method is found to be the best of all the techniques studied as it uses two different thresholds, a low threshold and a high threshold to detect strong and weak edges, thus eliminating the problem of inclusion false edges and discarding the valid edge points. We have used the Canny method in our work.

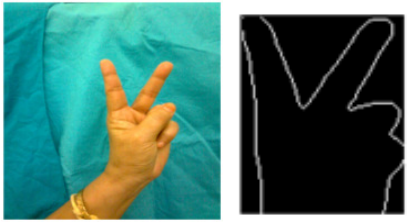

Fig. 7. Edge detction. (a) Input image, (b) Edge detected image

## D. Feature Extraction

Extraction of unique features for each gesture which are independent of human hand size and light illumination is important. We have used K-L transform as it provides advantages like it completely de-correlates the original data, the total entropy is minimized, and for any amount of compression the mean square error in the reconstruction is minimized. Some of the results are shown in fig. 8 which shows the image along with the Eigen vectors obtained using K-L Transform.

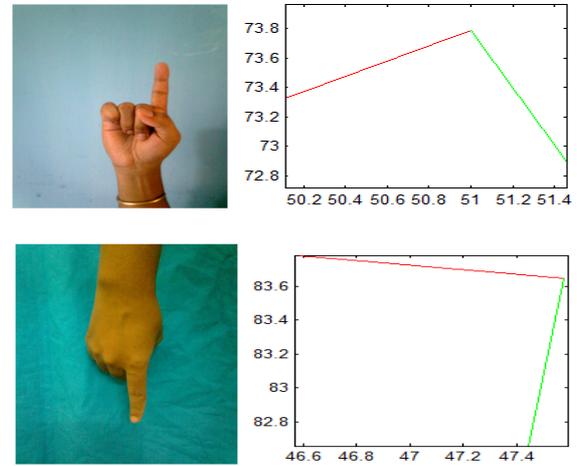

Fig 8. Features extracted for gesture 'UP' and 'DOWN'. (a) Input image, (b) Eigen vector plot.

## E. Classifier

We have used two types of classifier, one based on the angle made by the Eigen vector with reference axis and another based on Euclidean distance and made a comparative study between them and it was found that both of them gave the same results. The angle can be obtained considering either x or y-axis as reference axis but must be constant for all symbols. We have considered angles with respect to the x-axis. Table I shows the classification based on the angle between the Eigen vector and the x-axis. After rigorous experimentation it was found that for different images of a particular gesture the angle lie within certain limit. Based on that knowledge thresholds are set to classify the gesture. Here, we have considered only the angle made by the first Eigen vector, as the angle made by the second Eigen vector just differs by an angle of 90° with the first one. Considering only a single angle made by any one of the Eigen vector is sufficient for the purpose.

Table I: Classification based on angle made between eigen vector and the x-axis

| Threshold Angle made by Eigen vector | Symbol/Hand gesture |
|---|---|
| 22 – 30 | UP |
| (-7) – (-12) | DOWN |
| 79 – 81 | LEFT |
| (-100) – (-101) | RIGHT |
| (-5) – (-7) | VICTORY |
| (-0.6) – (-4) | THREE |
| 0 – 8 | FOUR |
| (-103) – (-106) | SMALL |
| (-44) – (-66.9) | THUMBS DOWN |
| (-67) – (-69) | LITTLE |

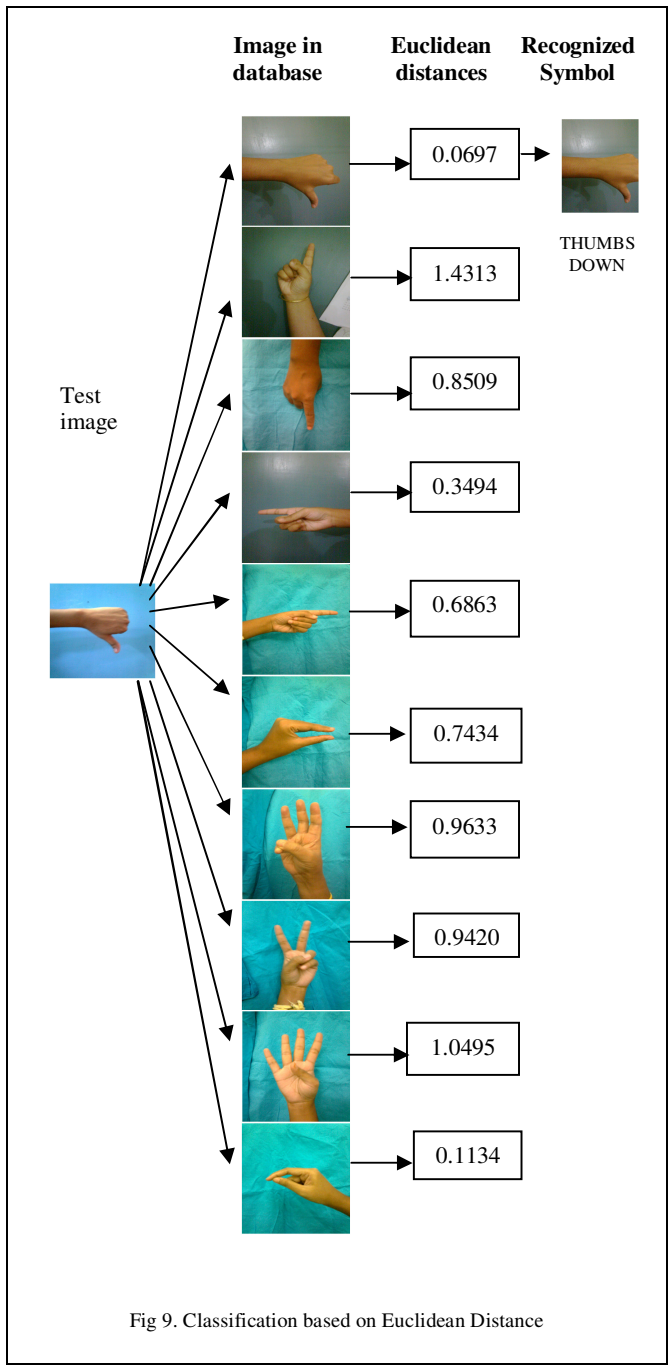

Fig 9. Classification based on Euclidean Distance

The second classifier is the classification based on the Euclidean distance in which Euclidean distance is found out between image to be tested and the images of each gesture already available in the database. Then the lowest distance value is determined from the set of values. The gesture to which the lowest distance indicates is the required gesture for the tested image. Fig. 9 shows the process by which the gesture is recognized by this type of classifier.

## 5. RESULTS AND DISCUSSIONS

We have taken 10 hand gesture images each having 10 samples, thus a total of 100 images were tested. Any arbitrary image after undergoing the proper steps was found to produce the following results. In table II, we made a comparative study on two classifiers, angle based and Euclidean distance based for few images and seen that both gave the same result. Thus, we arrive at a point that angle based classification is enough to classify different hand gestures.

From the above Table II, we conclude that almost all the 10 symbols were identified properly. Except some few gestures like FIVE, STOP, PHONE which is not shown here, their accuracy rate dropped down to 40%, 50% and 50% respectively.

Table II. Recognition rate obtained for different symbols

| Symbol | Total numbers | Correct numbers | Success rate |
|---|---|---|---|
| UP | 10 | 9 | 90% |
| DOWN | 10 | 10 | 100% |
| LEFT | 10 | 10 | 100% |
| RIGHT | 10 | 10 | 100% |
| VICTORY | 10 | 9 | 90% |
| THREE | 10 | 10 | 100% |
| FOUR | 10 | 9 | 90% |
| SMALL | 10 | 10 | 100% |
| THUMBS DOWN | 10 | 9 | 90% |
| LITTLE | 10 | 10 | 100% |

Here, in this paper we have shown the hand gesture recognition for 10 different gestures and for each gesture we have taken 10 samples. It has been classified based upon the angles made by the Eigen vectors with the axis (x-axis). A comparative study was made and concluded that the result was same even if it was classified based on Euclidean distance.

## 6. CONCLUSIONS

Here, we have used MATLAB version 7.6 (R2008a) as a software tool, and hardware used is Intel® Pentium® CPU B950 @ 2.10GHz processor machine, Windows 7 Home basic (64 bit), 4GB RAM and a 2 MP camera. In this paper, the steps that we have used for recognizing different hand gestures are skin filtering, edge detection, K-L transform and finally a proper classifier, where we have used angle based classification to detect which symbol the test image belongs to. The experiment was performed with bare hands while interacting with the computer. We have used Euclidean distance based classifier in order to compare the result from the angle based classifier and it was found that both give the same result. Thus accurate result was obtained. However, difficulty faced by us was it was not able to recognize gestures like STOP, FIVE whose result was similar to the FOUR and PHONE which was similar to THUMBS DOWN. We wish to overcome such difficulties in the near future. Moreover, here we have dealt with few gestures, so later we can extend to other gestures and implement it in different fields.

## REFERENCES

[1.] T. Kapuscinski and M. Wysocki, "Hand Gesture Recognition for Man-Machine interaction", *Second Workshop on Robot Motion and Control*, October 18-20, 2001, pp. 91-96.

[2.] C. Yu, X. Wang, H. Huang, J. Shen and K. Wu, "Vision-Based Hand Gesture Recognition Using Combinational Features", *IEEE Sixth International Conference on Intelligent Information Hiding and Multimedia Signal Processing,* 2010, pp. 543-546.

[3.] P.S. Rajam and G. Balakrishnan, "Real Time Indian Sign Language Recognition System to aid Deaf-dumb People", *IEEE*, 2011, pp. 737-742.

[4.] A. Malima, E. Ozgur and M. Cetin, "A Fast Algorithm for Vision-Based Hand Gesture Recognition for Robot Control", *IEEE*, 2006.

[5.] Manigandan M and I.M Jackin, "Wireless Vision based Mobile Robot control using Hand Gesture Recognition through Perceptual Color Space", *IEEE International Conference on Advances in Computer Engineering*, 2010, pp. 95-99.

[6.] D.Y. Huang, W.C. Hu and S.H. Chang, "Vision-based Hand Gesture Recognition Using PCA+Gabor Filters and SVM", *IEEE Fifth International Conference on Intelligent Information Hiding and Multimedia Signal Processing*, 2009, pp. 1-4.

[7.] E. Koh, J. Won and C. Bae, "On-premise Skin Color Modeing Method for Vision-based Hand Tracking", *The 13th IEEE International Symposium on Consumer Electronics (ISCE)*, 2009, pp. 908-909.

[8.] J.L. Raheja, K. Das and A. Chaudhury, "An Efficient Real Time Method of Fingertip Detection", *International Conference on Trends in Industrial Measurements and automation (TIMA)*, 2011, pp. 447-450.

[9.] J. Rekha, J. Bhattacharya and S. Majumder, "Shape, Texture and Local Movement Hand Gesture Features for Indian Sign Language Recognition", *IEEE*, 2011, pp. 30-35.

[10.] R. Gopalan and B. Dariush, "Towards a Vision Based Hand Gesture Interface for Robotic Grasping", *The IEEE/RSJ International Conference on Intelligent Robots and Systems*, October 11-15, 2009, St. Louis, USA, pp. 1452-1459.

[11.] Y. Fang, K. Wang, J. Cheng and H. Lu, "A Real-Time Hand Gesture Recognition Method", *IEEE ICME*, 2007, pp. 995-998.

[12.] P. Kumar, S.S. Rautaray and A. Agrawal, "Hand Data Glove: A New Generation Real-Time Mouse for Human-Computer Interaction", *IEEE 1st Int'l Conf. on Recent Advances in Information Technology (RAIT)*, 2012.

[13.] J.H. Kim, Nguyen Duc Thang and Tae-Seong Kim, "3-D Hand Motion Tracking and Gesture Recognition Using a Data Glove", *IEEE International Symposium on Industrial Electronics (ISlE)*, July 5-8, 2009, Seoul Olympic Parktel, Seoul , Korea, pp. 1013-1018.

[14.] M. Elmezain, A.A. Hamadi, G. Krell, S.E. Etriby and B. Michaelis, "Gesture Recognition for Alphabets from Hand Motion Trajectory Using Hidden Markov Models", IEEE International Symposium on Signal Processing and Information Technology, 2007, pp. 1192-1197.

[15.] J. Weissmann and R. Salomon, "Gesture Recognition for Virtual Reality Applications Using Data Gloves and Neural Networks", *IEEE*, 1999, pp. 2043-2046.

[16.] A. Cavallo, "Primitive Actions Extraction for a Human Hand by using SVD*", IEEE 9th International Symposium on Intelligent Systems and Informatics (SISY)*, September 8-10, 2011, Subotica, Serbia, pp. 413-419.

[17.] M.P. Paulraj, S. Yaacob, Mohd S.B.Z Azalan, R. Palaniappan, "A Phoneme Based Sign Language Recognition System Using Skin Color Segmentation", *IEEE 6th International Colloquium on Signal Processing & Its Applications (CSPA)*, 2010, pp. 86-90.

[18.] Y.F. Admasu and K. Raimond, "Ethiopian Sign Language Recognition Using Artificial Neural Network", *IEEE 10th International Conference on Intelligent System Design and Applications*, 2010, pp. 995-1000.

[19.] Stefan Oniga, János Vegh and Ioan Orha, "Intelligent Human-Machine Interface Using Hand Gestures Recognition", *IEEE*.

[20.] A.S. Ghotkar, R. Khatal, S. Khupase, S. Asati and M. Hadap, "Hand Gesture Recognition for Indian Sign Language", *IEEE International Conference on Computer Communication and Informatics (ICCCI)*, Jan. 10-12, 2012, Coimbatore, India.

[21.] L.K. Lee, S.Y. An, and S.Y. Oh, "Robust Fingertip Extraction with Improved Skin Color Segmentation for Finger Gesture Recognition in Human-Robot Interaction", *WCCI 2012 IEEE World Congress on Computational Intelligence*, June, 10-15, 2012, Brisbane, Australia.

[22.] R. Verma and A. Dev, "Vision based Hand Gesture Recognition Using Finite State Machines and Fuzzy Logic", *IEEE*, 2009.

[23.] J.B. Kim, K.H. Park**,** W.C. Bang and Z. Z. Bien, "Continuous Gesture Recognition System for Korean Sign Language based on Fuzzy Logic and Hidden Markov Model", *IEEE*, 2002, pp. 1574-1579.

[24.] Marcus V. Lamar, Md. Shoaib Bhuiyan, and Akira Iwata, "Hand Alphabet Recognition Using Morphological PCA and Neural Networks"*, IEEE*, 1999, pp. 2839-2844.

[25.] M.S. Chang and J.H. Chou, "A Friendly and Intelligent Human-Robot Interface System Based on Human Face and Hand Gesture", *IEEE/ASME International Conference on Advanced Intelligent Mechatronics,* Suntec Convention and Exhibition Center Singapore, July 14-17, 2009, pp. 1856-1861.

[26.] A.A. Singh, S. Banerjee and S. Chaudhury, "A Gesture Based Interface for Remote Robot Control", *IEEE*, 1998, pp.158-161.